\PassOptionsToPackage{table}{xcolor}
\documentclass[twocol]{ametsocV6.1}
\usepackage[absolute,overlay]{textpos}
\usepackage{hyperref}
\usepackage{amssymb} 
\usepackage{graphicx}
\usepackage{adjustbox}
\usepackage{booktabs, multirow} 
\usepackage{soul}
\usepackage{changepage,threeparttable} 
\usepackage{array}
\usepackage{makecell}

\newcolumntype{P}[1]{>{\raggedright\arraybackslash}p{#1}}

\definecolor{green1}{HTML}{407800}

\title{ANALYZING AND EXPLORING TRAINING RECIPES FOR LARGE-SCALE TRANSFORMER-BASED WEATHER PREDICTION}

\authors{Jared D. Willard,\aff{a}\correspondingauthor{Jared D. Willard, jared.d.willard@gmail.com}
Peter Harrington,\aff{a}
Shashank Subramanian,\aff{a}
Ankur Mahesh,\aff{b}
Travis A. O'Brien,\aff{c}
William D. Collins\aff{b,d}
}

\affiliation{\aff{a}{Lawrence Berkeley National Laboratory, National Energy Research Scientific Computing Center}\\
\aff{b}{University of California, Berkeley, Department of Earth and Planetary Science}\\
\aff{c} Indiana University Bloomington, Department of Earth and Atmospheric Sciences\\
\aff{d}{Lawrence Berkeley National Laboratory, Earth \& Environmental Sciences Area}}

\renewcommand{\abstract}[1]{}

\renewenvironment{abstract}{}{}
\begin{document}

\maketitle

\abstract{The rapid rise of deep learning (DL) in numerical weather prediction (NWP) has led to a proliferation of models which forecast atmospheric variables with comparable or superior skill than traditional physics-based NWP. However, among these leading DL models, there is a wide variance in both the training settings and architecture used. Further, the lack of thorough ablation studies makes it hard to discern which components are most critical to success. In this work, we show that it is possible to attain high forecast skill even with relatively off-the-shelf architectures , simple training procedures, and moderate compute budgets. Specifically, we train a minimally modified SwinV2 transformer on ERA5 data, and find that it attains superior forecast skill when compared against IFS. We present some ablations on key aspects of the training pipeline, exploring different loss functions, model sizes and depths, and multi-step fine-tuning to investigate their effect. We also examine the model performance with metrics beyond the typical ACC and RMSE, and investigate how the performance scales with model size.}
%
%
%
%
%
%
\section{INTRODUCTION}

\begin{figure*}
    \centering
    \begin{adjustbox}{center}
    \includegraphics[width=1.2\textwidth]{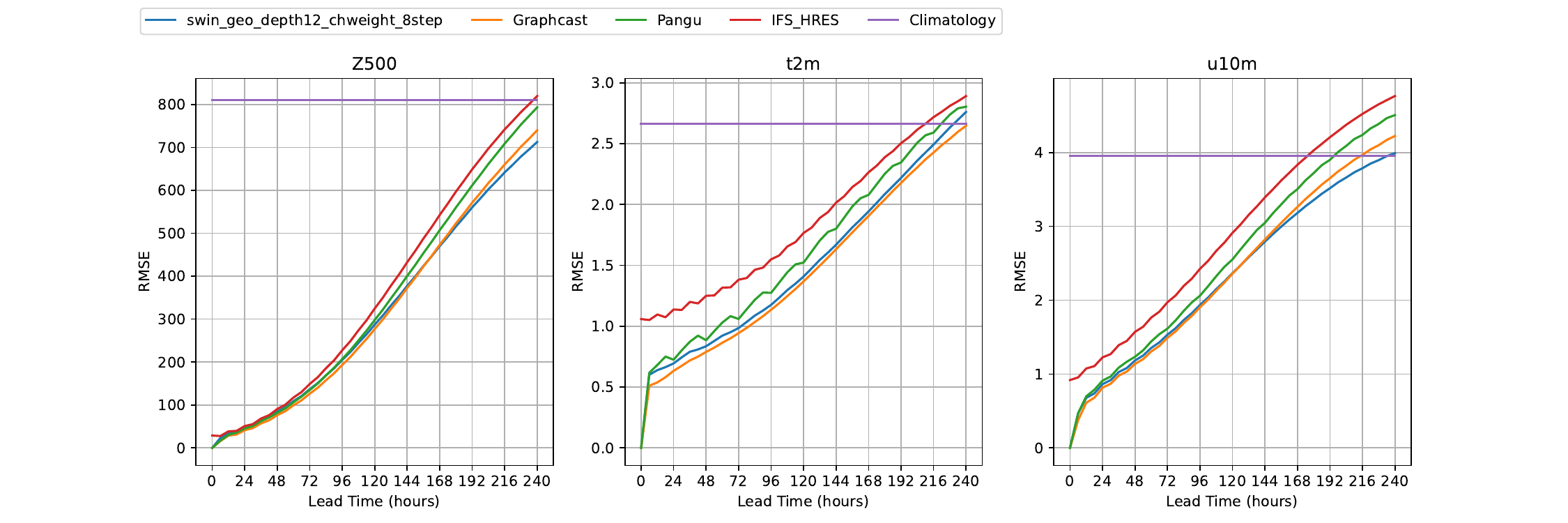}
    \end{adjustbox}
    \caption{Weatherbench 2 deterministic RMSE comparison of forecasts of z500, t2m, and u10m at lead times up to 10 days for the swin model using channel-weighting, 8-step fine-tuning, and latitude-weighted loss, IFS\_HRES, Pangu-Weather, and Graphcast compared to climatology.}
    \label{fig:wb2_score}
\end{figure*}
As the impacts of climate change continue to grow in severity, it is becoming more important than ever to forecast weather phenomena with high accuracy and fidelity. While operational forecasting has long been performed by numerical weather prediction (NWP) models like the Integrated Forecast System (IFS), the availability of high-quality reanalysis data \citep{hersbach2020era5} and the onset of advanced deep learning techniques has led to a proliferation of data-driven forecast models in recent years. The power of deep learning in weather prediction has progressed rapidly, producing models that compete with or outperform leading NWP systems in key forecast metrics \citep{lam2023learning,bi2023accurate} only 4 years after the earliest pioneering works \citep[e.g.,][]{gmd-11-3999-2018}. This, combined with the fact that deep learning weather models offer unique capabilities that can augment the capabilities of NWP \citep{ben2023rise}, has generated substantial interest in how to design the most effective deep learning approaches for weather prediction.

Recent literature has proposed an abundance of training recipes, network architectures, inference configurations, and compute budgets for the task of global medium-range forecasting. Proposed architectures range from graph neural networks \citep{keisler2022forecasting,lam2023learning}, transformers \citep{bi2023accurate, chen2023fengwu, chen2023fuxi, nguyen2023scaling}, neural operators \citep{pathak2022fourcastnet, bonev2023spherical}, and convolutional neural networks \citep{karlbauer2023advancing}; the training recipes include different loss functions, normalization methods, time-stepping schemes, and varied sets of atmospheric variables and resolutions/grids. Expanding beyond just deterministic forecasting and conventional deep learning, other works have also explored diffusion models \citep{price2023gencast} and hybrid physics-ML schemes \citep{kochkov2023neural}. This progress, while exciting, also presents a challenge for researchers, as the differences between models are multifold and sometimes entangled, so separating their effects is not always possible. While some works present ablations and analysis on parts of their models, it is challenging to do so comprehensively, and there remains a need for analysis under restricted and controlled settings.

With the advent of open benchmarks providing detailed and informative model evaluation \citep{rasp2023weatherbench, brenowitz2024practical}, comparisons between and within models are becoming easier, and the recent Stormer model \citep{nguyen2023scaling} exemplifies what can be done with extensive ablation studies. Their work takes a standard Vision Transformer (ViT) and explores the impact of key components of the training pipeline and model architecture on the downstream performance of the model, assessed by the root-mean-square-error (RMSE) of deterministic forecasts for key weather variables. Their improvements achieve highly competitive accuracy when compared against other leading models at $\sim1.4^\circ$ resolution.

In this work, we take a similar approach and aim to assess the effect of different training settings, and the interplay between them, but \emph{at full ERA5 resolution} ($0.25^\circ$). The decision to work at full resolution is motivated by the fact that high-resolution forecasts are simply more useful, and that excessive blurring and lack of fine-scale detail is a current shortcoming of current deep-learning-based models \citep{ben2023rise, price2023gencast, brenowitz2024practical}. We choose the SwinV2 \citep{liu2022swin} architecture as a representative transformer-based architecture that works well at high resolutions, and is a relatively ``off-the-shelf'' architecture widely available in deep learning libraries. With SwinV2, we explore several different aspects of the training pipeline. We do not aim to provide a comprehensive analysis of all proposed training strategies, but focus on what can achieve good deterministic skill at full ERA5 resolution on 1-5 day lead times. This focus is partly motivated by observations that multi-step fine-tuning and other methods that most improve skill at long lead times exacerbate the issue of blurring, as model predictions tend to resemble more of an ensemble mean rather than a deterministic forecast \citep{price2023gencast, brenowitz2024practical}. Our contributions are as follows:

\begin{itemize}

    \item We present a minimally modified SwinV2 model trained on ERA5 at full 0.25$^\circ$ resolution that outperforms IFS in deterministic skill.
    \item We present detailed ablations over key training and model settings, finding the effects of latitude-weighting, channel-weighting, and multi-step fine-tuning to be somewhat entangled; they are generally positive but not always constructive or additive with each other.
    \item We confirm that multi-step fine-tuning can improve RMSE but affect sharpness and ensemble spread in transformer architectures as well.

\end{itemize}

\section{DATASET \& MODEL DETAILS}

\subsection{Data}
We use the ERA5~\citep{hersbach2020era5} dataset, provided by ECMWF (European
Center for Medium-Range Weather Forecasting). ERA5 contains hourly reanalysis data at a spatial resolution of $0.25^{\circ}$ ($\sim25$ km) from years 1979 to present day. For this study we subsample ERA5 on $\Delta t = 6$ hour time intervals, and select 73 variables from the full dataset to include in the model (this closely follows previous work, e.g. \citet{nguyen2023scaling,bi2022pangu}). These are geopotential height (z), winds (u, v), temperature (t), and specific humidity (q) at 13 vertical pressure levels (50hPa, 100hPa, 150hPa, 200hPa, 250hPa, 300hPa, 400hPa, 500hPa, 600hPa, 700hPa, 850hPa, 925hPa, and 1000hPa), along with 8 single-level/surface variables: surface winds at 10m and 100m (u10, v10, u100, v100), 2m temperature (t2m), surface pressure (sp), mean sea level pressure (msl), and total column water vapor (tcwv). We also include as static additional inputs the land-sea mask, orography, and cosine of zenith angle (indicating time of day/year). As mentioned in the previous section, we focus on results at full ERA5 resolution and thus do not downsample the data as in \cite{nguyen2023scaling}. All data is normalized by the global mean and standard deviation per variable before training. We use years 1979-2015 as well as 2019 for training data, and 2016-17 for validation, then evaluate on 2018 in line with other recent DL-based forecast models.

\subsection{Model Architecture}

We base our model on the SwinV2~\citep{liu2022swin} implementation available in v0.9.2 of the \texttt{timm} library\footnote{\href{https://github.com/huggingface/pytorch-image-models}{https://github.com/huggingface/pytorch-image-models}}. While other DL-based forecast models have made extensive modifications to Swin backbones \citep{bi2023accurate, chen2023fuxi}, we find it sufficient to just minimally modify two aspects of the SwinV2 architecture:
\begin{itemize}
    \item \textbf{Window shifting:} The shifting of attention windows in Swin is implemented as a \texttt{torch.roll} operation, followed by masking to ensure attention within shifted windows doesn't cross image boundaries (since 2D images are non-periodic generally). In the case of ERA5, rolling along the horizontal (zonal) direction is fine, since this axis is periodic, so we only apply the masking along the vertical dimension. This slightly simplifies the SwinV2 code.
    \item \textbf{Position embedding:} In SwinV2 the position biases within attention windows are carefully defined in a relative coordinate system, to better allow transferring across multiple resolutions/window sizes. In our case we are restricting ourselves to data at fixed resolution, and thus find it sufficient to drop the relative position embeddings in favor of a standard ``absolute'' position embedding as in a standard ViT, which is added to the latent space immediately after patch embedding.
    \item \textbf{Non-hierarchical structure:} SwinV2 uses a common vision transformer technique to incorporate a hierarchy across the network layers that sequentially merges patches and decreases resolution in order to both reduce computational cost and model various scales. We alter the model to be non-hierarchical and maintain the same feature resolution in all layers, which has shown to be effective for spatiotemporal forecasting in earth science \citep{gao2022earthformer}. 
\end{itemize}

With these modifications we train our forecast models in the standard setting, giving the state $\mathbf{X}_t$ as input and predicting the next state $\mathbf{X}_{t+\Delta t}$. In inference, the model is rolled out autoregressively to produce forecasts for lead times larger than $\Delta t$. For our baseline SwinV2 model we use an embedding dimension of 768, depth of 12 layers, patch size of 4, 8 attention heads, and local attention window size 9x18. The model is trained with latitude-weighted (described in the following section) $\mathcal{L}_2$ loss using the Adam optimizer with a DropPath rate of 0.1, learning rate of 0.001, and batch size of 64, and trains for 70 epochs on 64 A100 GPUs. Activation checkpointing is used as necessary to fit fine-tuning configurations into GPU memory. Modifications and ablations with respect to this baseline model are described in subsequent sections.

\subsection{Ablations \& Experiments}

\textbf{Model size:} In preliminary experimentation we found that, for a given embedding dimension and depth, the largest possible window size and smallest patch sizes performed best. Thus to probe the effect of larger model sizes, we explore growing both the depth and width dimensions. We evaluate the performance of a model with twice as many layers (\texttt{depth=24}), as well as double the embedding dimension (\texttt{embed\_dim}=1536). 

\textbf{Channel weighting:} Recent work has found it beneficial to carefully weight channels (different weather variables at different vertical levels) in the loss function during training \citep{nguyen2023scaling}. In particular, the method first pioneered by \cite{lam2023learning} of down-weighting as pressure level decreases (higher vertical levels are weighted less) and down-weighting according to the standard deviation of temporal differences ($\sigma_{\mathbf{\delta X}}$) has been found to work well. Beyond these physically-motivated weights, \cite{lam2023learning} also manually impose weights that preferentially emphasize certain surface variables, like t2m, which we adopt as well for consistency. Similar to \cite{nguyen2023scaling} we evaluate the effect of this configuration compared against the standard loss where all channels are given equal weight.

We note this experiment is also entangled with ``direct'' vs. ``residual'' prediction -- the channel-weighting method, which partially weights according to the temporal differences ($\sigma_{\mathbf{\delta X}}$), predicts the difference between the input and target timesteps, whereas the standard $\mathcal{L}_2$ loss directly predicts the target timestep. In this work we do not explore disentangling these two configurations, but in principle one could separately apply pressure-level/$\sigma_{\mathbf{\delta X}}$ and direct/residual prediction.

\textbf{Multi-step fine-tuning:} Implemented in many works, \citep{pathak2022fourcastnet, lam2023learning, bonev2023spherical, nguyen2023scaling, chen2023fuxi}, this method aims to improve long-term forecast performance by optimizing the loss over multiple (autoregressive) timesteps. While this improves deterministic RMSE at longer lead times, it has also been found to adversely affect forecast sharpness and ensemble spread \citep{price2023gencast, brenowitz2024practical}. We evaluate the effects of fine-tuning models with 4 and 8 timesteps to build upon the observations found in previous works. Fine-tuning is done at a reduced learning rate of 1e-4 for 15 epochs. 

\textbf{Latitude-weighting:} A large number of works, dating back to the original WeatherBench baseline \cite{rasp2020weatherbench}, have additionally weighted the loss according to the cosine of each grid cell's latitude. This is motivated by the spherical geometry, and compensates for the difference in area between cells near the poles versus the equator in the equiangular projection. We examine the performance of models with and without this weighting applied, though it is applied by default if unspecified.


\subsection{Evaluation}

We use the recent open-source \texttt{earth2mip} package\footnote{\href{https://github.com/NVIDIA/earth2mip}{https://github.com/NVIDIA/earth2mip}} \citep{brenowitz2024practical} to score and evaluate models. Scores are averaged over 11 initial conditions evenly spaced throughout 2018, and forecasts are rolled out to 7 days at 6 hour intervals. We primarily focus on latitude-weighted deterministic RMSE to compare between different models, but doing so imposes trade-offs with other metrics and desirable aspects of forecast quality (e.g. sharpness/bluriness). Thus we additionally measure energy spectra and the ensemble spread/skill in a lagged-ensemble \citep{tracton1993operational} forecast for some of our models, to further illuminate these trade-offs.

\section{RESULTS}
\subsection{Model size, channel weighting, \& multi-step fine-tuning}
\begin{figure*}[h]
    \centering
    \includegraphics[width=\textwidth]{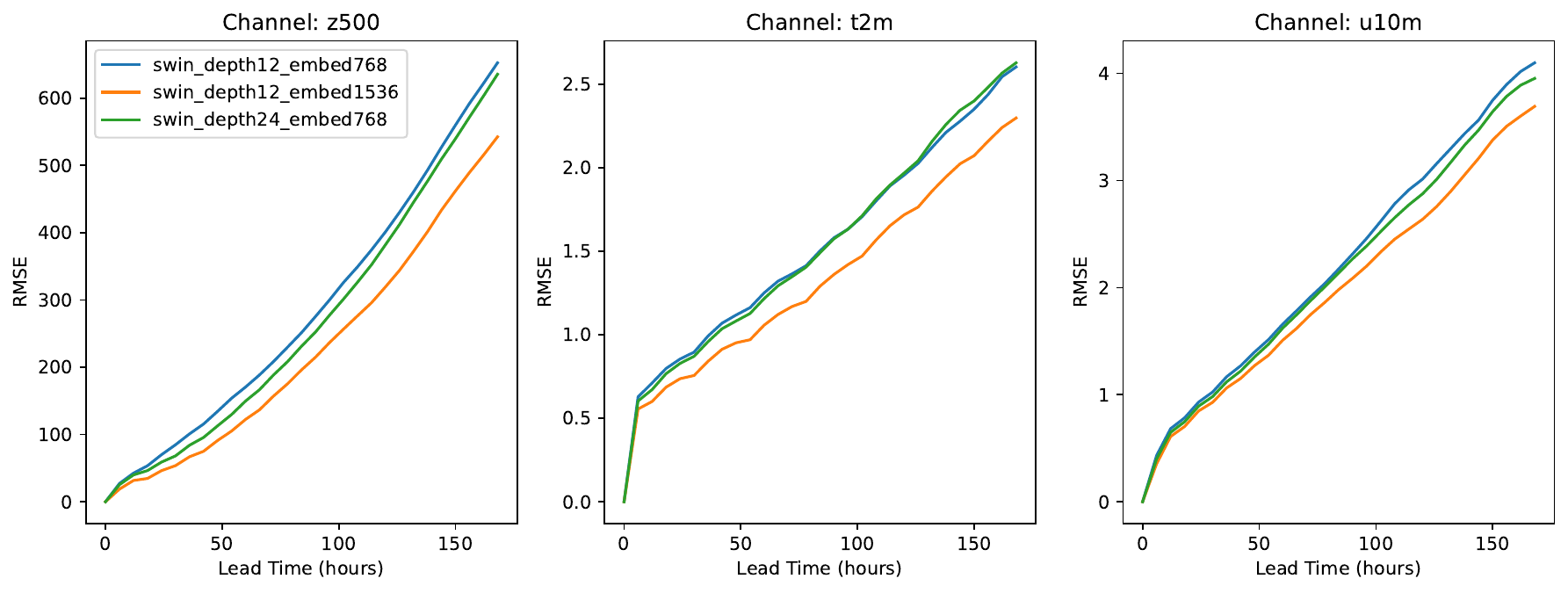}
    \caption{RMSE comparison of forecasts at lead times up to 7 days for the different model depth and embedding dimension for the SwinV2 model}
    \label{fig:model_size_lineplot}
\end{figure*}

We examine variations of model size and the effect on forecast skill in Figure \ref{fig:model_size_lineplot}. While doubling the embedding dimension or depth improves performance generally, the model variant with increased embedding dimension consistently outperforms the deeper model at all lead times. The deeper variant struggles to beat the baseline in t2m and u10, and the gap between it and the model with larger embedding dimension becomes as large as 10-15\% in RMSE at 7-day lead times.


\begin{figure*}[h]
    \centering
    \includegraphics[width=\textwidth]{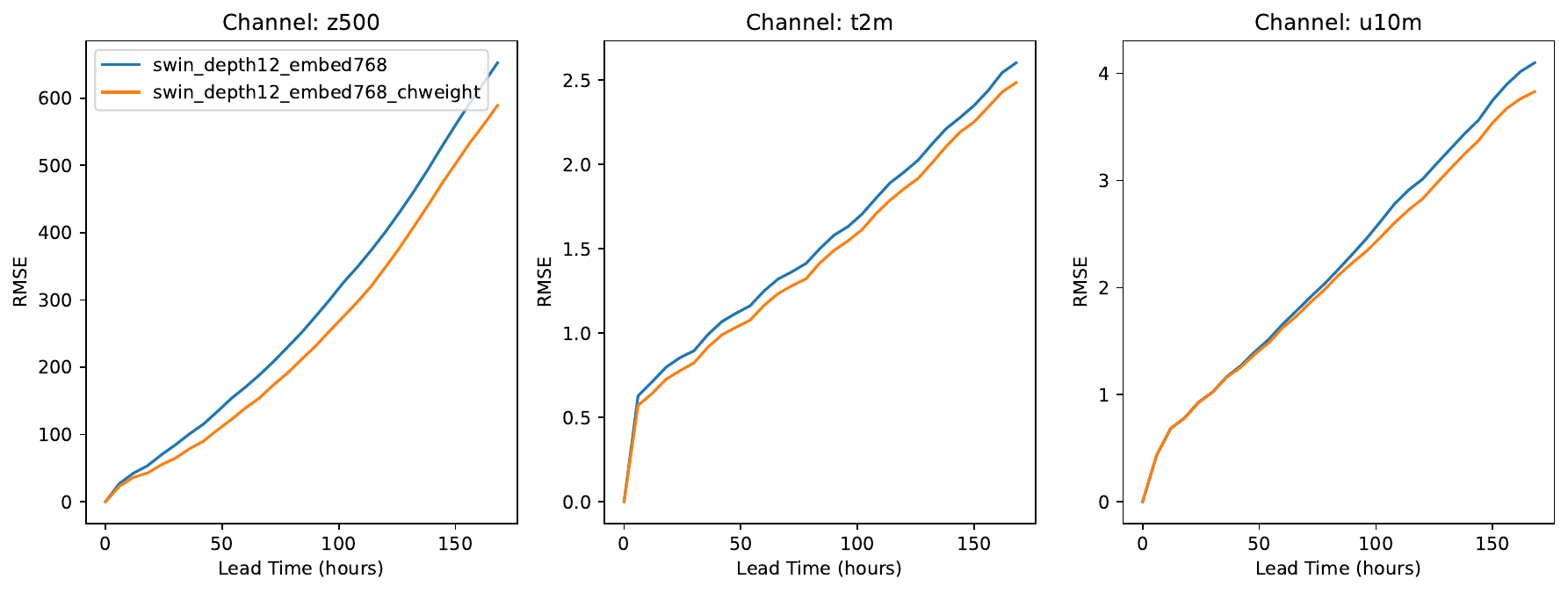}
    \caption{RMSE comparison of forecasts at lead times up to 7 days for the baseline depth 12 and embedding dimension 768 SwinV2 model with and without custom channel-weighting}
    \label{fig:chw_line_plot}
\end{figure*}

In Figure~\ref{fig:chw_line_plot} we compare the baseline configuration against a model trained with channel-weighted loss and confirm that the channel-weighting seems to improve the model's forecasting accuracy across the majority of lead times. For some variables (e.g., u10) the improvement is more apparent at later lead times. We note the improvement is present even in variables which are actually down-weighted by the loss (e.g., z500, which is weighted less since it is further from the surface).


\begin{figure*}
    \centering
    \includegraphics[width=\textwidth]{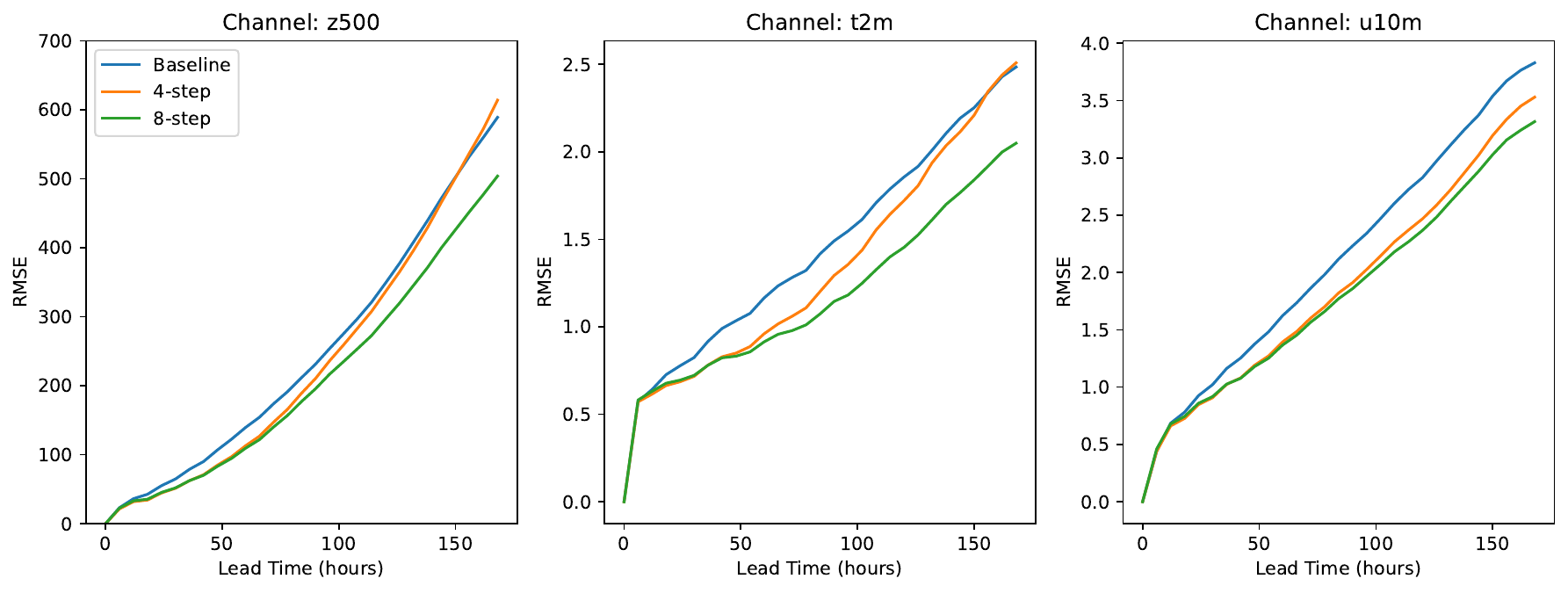}
    \caption{RMSE comparison of forecasts at lead times up to 7 days for the model trained with custom channel-weighting, depth 12, and embedding dimension 768 alongside its variants with 4 and 8-step fine tuning}
    \label{fig:nstep_line_plot}
\end{figure*}

In Figure \ref{fig:nstep_line_plot} we show the effect of multi-step fine-tuning and channel-weighting simultaneously applied, finding it effective in improving RMSE as previous works have shown. The overall effect seems to be that multi-step fine-tuning improves performance most at lead times much larger than the fine-tuning window. For example, up to lead times of $\sim$2-3 days, there is no benefit from fine-tuning up to 8 steps (48 hours) versus just 4 (24 hours) -- both improve over the baseline by the same amount. However, at lead times of 5 days and beyond, the gap between the two fine-tuning configurations has increased substantially and the 8-step model performs much better.


\subsection{Downstream effects of multi-step fine-tuning}

\begin{figure*}
    \centering
    \includegraphics[width=\textwidth]{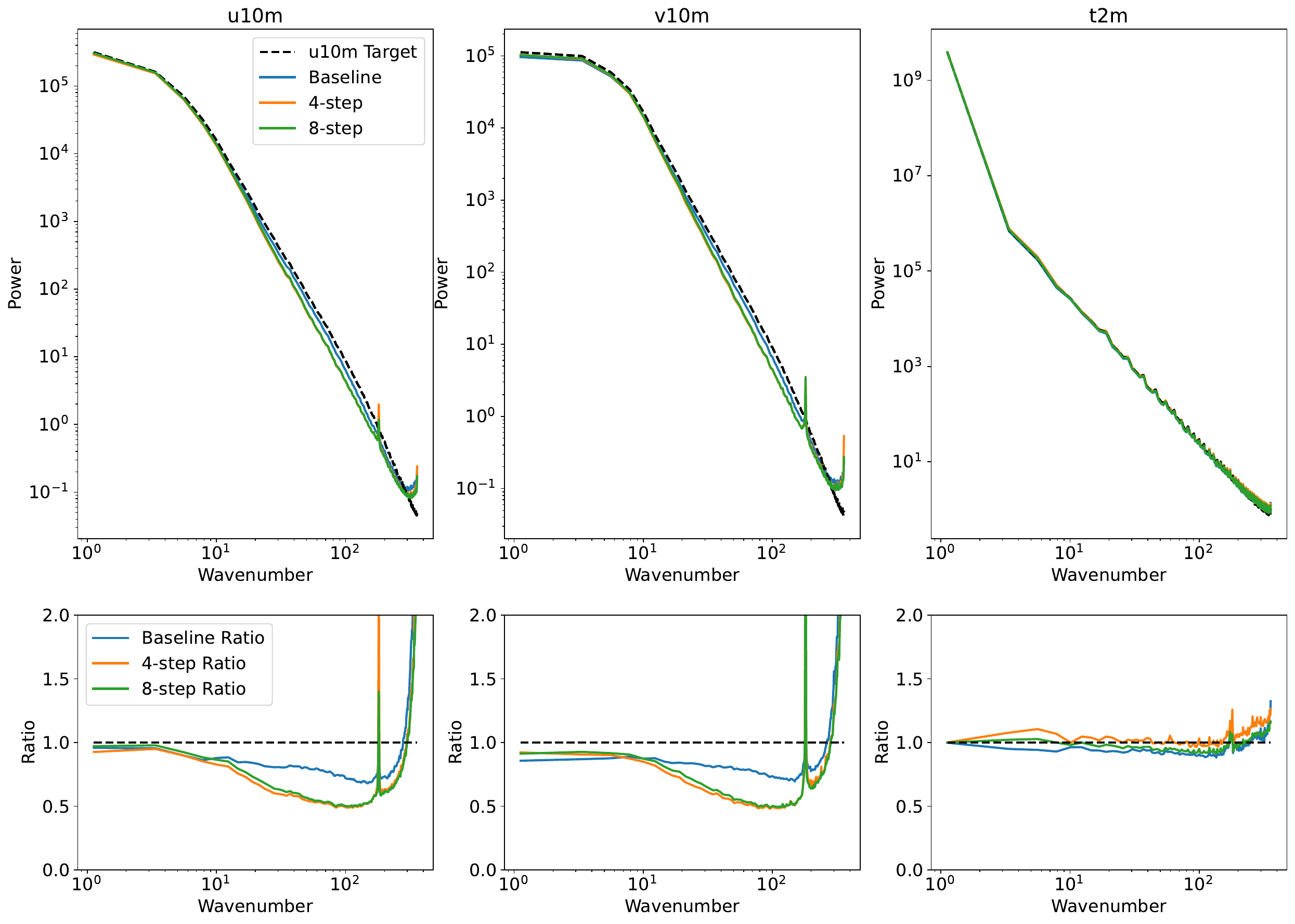}
    \caption{Spatial frequency representation for the z500, t2m, and u10m variables across three model configurations: the baseline model using channel-weighting, and two fine-tuned versions of that model using 4-step and 8-step fine-tuning respectively. For each variable and model, the upper plot shows the power spectral density (PS1D) of both the target ERA5 (black line) and prediction (red line) on a logarithmic scale. The lower plot displays the ratio of predicted to actual PS1D values, with a ratio of 1 (dashed line) indicating perfect alignment. Ratios above or below 1 indicate overestimations or underestimations of power at specific spatial frequencies, respectively.}
    \label{fig:psd_3panel}
\end{figure*}

Given the stark improvement in RMSE offered by multi-step training, it is worth reiterating that these improvements come at the cost of other qualities desirable in weather models: forecast sharpness and ensemble spread. As a quick demonstration, we show in Figure \ref{fig:psd_3panel} the power spectra of model predictions compared against the ground truth ERA5 (averaged over all lead times and initial conditions). The multi-step fine-tuned models are clearly deficient in higher wavenumbers for u10, indicating blurring. This effect is not present in all variables, as seen in the spectra for, e.g. surface temperature, whose fine-scale power might be more dominated by (static) orographic features like mountain ranges and coastlines rather than dynamics. The spikes and pileup near Nyquist frequency are caused by the patch embedding in SwinV2.

\begin{figure*}
    \centering
    \includegraphics[width=\textwidth]{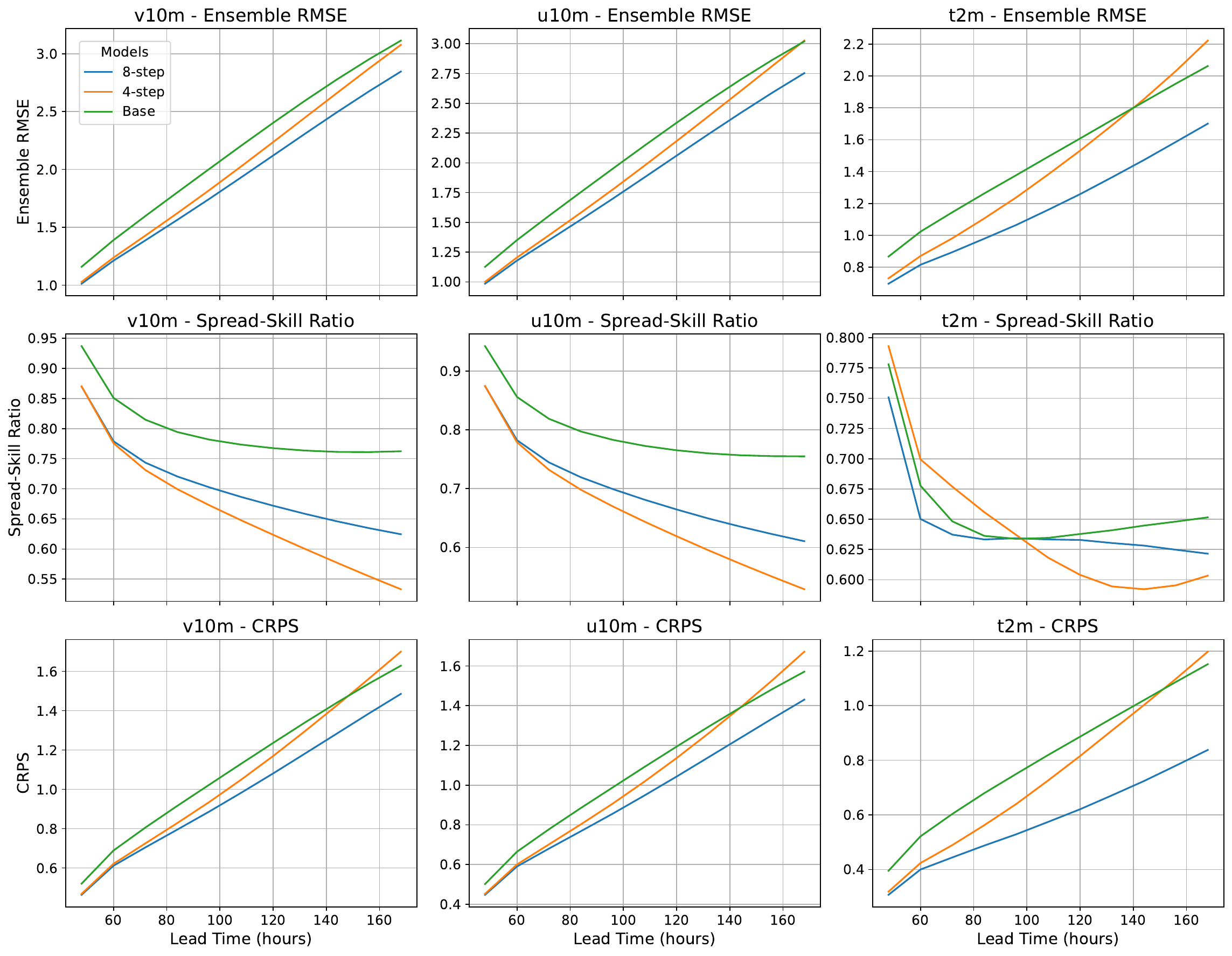}
    \caption{Lagged ensemble forecast performance for v10, u10, and t2m across three model configurations: the baseline model using channel-weighting, and two fine-tuned versions of that model using 4-step and 8-step fine-tuning. The first row is ensemble mean RMSE calculated over the 9 member ensemble, the second row is the spread skill calculated by dividing the standard deviations of predictions across the ensemble by the ensemble RMSE, and the third row is continuous ranked probability score calculated across the ensemble.}
    \label{fig:lagged_ens_9panel}
\end{figure*}

We examine spread and performance of an ensemble constructed from lagged forecasts in Figure \ref{fig:lagged_ens_9panel}. The lagged ensemble procedure and motivation is described in detail by \cite{brenowitz2024practical}, but generally we can expect that models which are more intrinsically more dispersive (i.e., create ensembles with larger spread) to have a larger spread/skill ratio, which should ideally be 1 for an optimal real-world ensemble. In Figure \ref{fig:lagged_ens_9panel} we observe that indeed the spread-skill ratio is diminished for both multi-step fine-tuned models, confirming the issues presented in \cite{brenowitz2024practical}. This decrease in spread-skill ratio is not catastrophic, as there still appears to be ensemble skill gains from fine-tuning as shown in ensemble mean RMSE and CRPS in the first and third rows respectively. In particular, the 8-step fine-tuned model has better skill at all lead times than the baseline for all three variables in both metrics. 

\subsection{Effects of latitude-weighted loss}
Though the previous results have all incorporated the latitude-weighted loss, in Table \ref{tab:area_weight_results} we examine this choice more closely and compare across several model configurations with and without applying latitude-weighting. The results are mixed, highlighting the need for caution when drawing conclusions from just one or two experiments alone. For the 8-step fine-tuned model, the latitude-weighted loss is generally helpful and achieves the lowest RMSE at lead times of 2 and 4 days (at 7-days the RMSE is roughly the same with and without latitude-weighting), with the best overall configuration using both channel-weighting and latitude-weighting in the loss. Surprisingly, the conclusion is reversed for a model that is only trained over single-step predictions -- the latitude-weighted loss does significantly poorer at all lead times, and the best configuration uses neither channel-weighting nor latitude-weighting. Thus the effects of channel-weighting, latitude-weighting, and multi-step training are entangled; since it is common practice to do more hyperparameter tuning using less expensive configurations first (e.g., tune 1-step training first and then apply multi-step training) this can cause problems during the model development process.

\scalebox{0.94}{\begin{threeparttable}[h]
\caption{RMSE  for Z500 (m) at 2,4,and 7 day forecast lead times for different model configurations trained with and without latitude-weighted loss. All models have depth 12 and 768 embedding dimension. The best results for 1-step and 8-step training are highlighted in bold.}\label{tab:area_weight_results}
\scriptsize
\begin{tabular}{@{}lcccccc@{}}\toprule

\makecell{Channel\\weighting} & \makecell{\# step\\training} & \makecell{Latitude-\\weighted} & \makecell{Z500\\(2day)} & \makecell{Z500\\(4day)} & \makecell{Z500\\(7day)} \\ 
\cmidrule{1-6}
 -- & 1 & -- & \bf{95.74} & \bf{237.36} & \bf{549.18} \\
 -- & 1 & \checkmark & 134.29 & 299.96 & 652.52 \\
 \checkmark & 1 & -- & 102.31  & 261.08 & 572.33 \\
 \checkmark & 1 & \checkmark & 106.94 & 253.72 & 589.08 \\
\cmidrule{1-6}
  -- & 8 & -- & 85.48 & 218.65 & \bf{503.28} \\
  -- & 8 & \checkmark & 84.01 & 217.19 & 507.65 \\
  \checkmark & 8 & -- & 88.48 & 227.77 & 505.18 \\
  \checkmark & 8 & \checkmark & \bf{83.20} & \bf{216.66} & 503.93  \\
\bottomrule
\\
\end{tabular}
\end{threeparttable}
}

\subsection{Additional experiments}

Beyond this main set of experiments, we also partially explore other methods proposed in the literature, but do not run complete ablation suites due to the initially poor performance we observe. In particular, we attempt to use the variable tokenization embedding \citep{nguyen2023scaling}, which uses a cross-attention operation to fuse information between variables after patch embedding. This had previously only been demonstrated on much coarser resolution, and we find that applying it to full 0.25$^\circ$ resolution data poses significant computational challenges due to the memory cost. With activation checkpointing and breaking the operation to run in chunks sequentially, we are able to get it to fit on 80GB A100s, but we observe a slight degradation in the training and validation loss, contrary to the observations of \citep{nguyen2023scaling}. Since the checkpointing and chunked computation significantly increases the runtime, we do not pursue this further.

We also explore the alternate negative log likelihood (NLL) loss from \citep{chen2023fengwu}, which down-weights the loss according to a network-predicted uncertainty. Initial experiments under this configuration showed the 1-step RMSE on the validation set lagging behind the baseline configuration, and the predictions were noticeably more blurry. Thus we did not find this configuration worth including in more extensive ablations.

\subsection{Weatherbench 2 Evaluation}
We evaluate one of our best performing SwinV2 models on Weatherbench 2 \citep{rasp2023weatherbench}, a widely used benchmark for data-driven weather forecasting. Using the deterministic evaluation option, we compare our swin model using channel-weighting, 8-step fine-tuning, and latitude-weighted loss against the IFS\_HRES \citep{ifs_hres_dataset}, Pangu-Weather \citep{bi2022pangu}, and Graphcast \citep{lam2023learning} for forecasts up to 10 days out at 6 hour intervals. To match existing Weatherbench 2 scores, we expanded our number of initial times to every 12 hour interval in the 2018 evaluation year (731 total).  The results are shown in Figure \ref{fig:wb2_score}. We see that for z500 the Swinv2 model generally outperforms IFS and is on par with Pangu-Weather but slightly worse than Graphcast for lead times less than 7 days, but surpasses both models beyond 7 days. For t2m, the SwinV2 model generally outperformed Pangu-Weather and IFS\_HRES but was slightly worse than Graphcast for all lead times. Lastly for u10m, the model generally outperformed Pangu-Weather at almost all lead times, and was similar to Graphcast but slightly worse until about 5 days lead times, beyond which the SwinV2 was more accurate in terms of RMSE. Overall  our SwinV2 model showed strong performance comparable with existing data-driven models that outperform IFS\_HRES, especially for forecasts extending beyond 5 to 7 days for z500 and u10m.



\section{Conclusions}
In this work we have demonstrated that relatively off-the-shelf architectures can outperform IFS and achieve highly competitive forecast skill with the proper training procedure. We find that increasing model size, applying channel weighting in the loss, and training over multiple time steps all improve deterministic forecast skill. We also see the effectiveness of the latitude-weighted loss to vary across different configurations, finding it more effective when used in conjunction with multi-step training. We confirm that multi-step training can also adversely affect forecast sharpness and ensemble spread, highlighting the need for other methods to stabilize rollouts and improve deterministic skill. We find some other innovations proposed in previous literature to either be ineffective in our setting or infeasible at full ERA5 resolution. Since our models are trained with moderate compute budgets (e.g., a total of 48 hours on 64 A100 GPUs for pre-training and fine-tuning the base model), we hope these models and findings will be of great practical use to the community.

\clearpage
\acknowledgments{
This research was supported by the Director, Office of Science, Office of Biological and Environmental Research of the U.S. Department of Energy under Contract No. DE-AC02-05CH11231 and by the Regional and Global Model Analysis Program area within the Earth and Environmental Systems Modeling Program. The research used resources of the National Energy Research Scientific Computing Center (NERSC), also supported by the Office of Science of the U.S. Department of Energy, under Contract No. DE-AC02-05CH11231. The computation for this paper was supported in part by the DOE Advanced Scientific Computing Research (ASCR) Leadership Computing Challenge (ALCC) 2023-2024 award `Huge Ensembles of Weather Extremes using the Fourier Forecasting Neural Network' to William Collins (LBNL).}

%
%
\subsection*{Code and Data Availability Statement.} 
The primary code used in this is study is available at \url{https://github.com/NERSC/swin_v2_weather/}. The repository README also has links to ERA5 data, neural network model weights, and precomputed statistics for normalization. The version of Earth2Mip used that contains the swin transformer implementation can be found at \url{https://github.com/jdwillard19/earth2mip-swin-fork}. 

%






%



\bibliographystyle{ametsocV6}
\bibliography{references}

\end{document}